\documentclass[10pt,twocolumn,letterpaper]{article}

\usepackage{cvpr}              
\definecolor{cvprblue}{rgb}{0.21,0.49,0.74}
\usepackage[pagebackref,breaklinks,colorlinks,allcolors=cvprblue]{hyperref}
\usepackage{pifont}
\usepackage{multirow}
\usepackage{graphicx}
\usepackage{array}
\usepackage{caption}
\usepackage{float}
\usepackage{listings}
\usepackage[accsupp]{axessibility} 
\usepackage[table]{xcolor}
\definecolor{paleblue}{HTML}{ECEDF6}

\floatstyle{plain}
\newfloat{listing}{t}{lop}
\floatname{listing}{Listing}

\def\paperID{9482} 
\def\confName{CVPR}
\def\confYear{2026}

\title{Guiding a Diffusion Model by Swapping Its Tokens}

\author{Weijia Zhang$^{1}$ ~~~~~~~ Yuehao Liu$^{1}$ ~~~~~~~ Shanyan Guan$^{2}$ ~~~~~~~ Wu Ran$^{1}$  \\ ~~~~~~~ Yanhao Ge$^{2}$~~~~~~~ Wei Li$^{2}$ ~~~~~~~ Chao Ma$^{1,}$\thanks{ Corresponding author.}\\
$^{1}$ MoE Key Lab of Artificial Intelligence, AI Institute, Shanghai Jiao Tong University \\ $^{2}$ vivo Mobile Communication Co., Ltd. \\
{\tt\small \{weijia.zhang, yuehao.liu, chaoma\}@sjtu.edu.cn} \\  
{\tt\small  \{guanshanyan, halege, liwei.yxgh\}@vivo.com}
}

\begin{document}
\maketitle
\begin{abstract}
Classifier-Free Guidance (CFG) is a widely used inference-time technique to boost the image quality of diffusion models. Yet, its reliance on text conditions prevents its use in unconditional generation. 
We propose a simple method to enable CFG-like guidance for both conditional and unconditional generation. The key idea is to generate a perturbed prediction via simple token swap operations, and use the direction between it and the clean prediction to steer sampling towards higher-fidelity distributions. In practice, we swap pairs of most semantically dissimilar token latents in either spatial or channel dimensions.
Unlike existing methods that apply perturbation in a global or less constrained manner, our approach selectively exchanges and recomposes token latents, allowing finer control over perturbation and its influence on generated samples.
Experiments on MS-COCO 2014, MS-COCO 2017, and ImageNet datasets demonstrate that the proposed Self-Swap Guidance (SSG), when applied to popular diffusion models, outperforms previous condition-free methods in image fidelity and prompt alignment under different set-ups. Its fine-grained perturbation granularity also improves robustness, reducing side-effects across a wider range of perturbation strengths.
Overall, SSG extends CFG to a broader scope of applications including both conditional and unconditional generation, and can be readily inserted into any diffusion model as a plug-in to gain immediate improvements. 
The code is available at \url{https://github.com/VISION-SJTU/SSG}.
\end{abstract}
\section{Introduction}
\label{sec:intro}

\begin{figure}[t]
    \centering
    \includegraphics[width=0.49\textwidth]{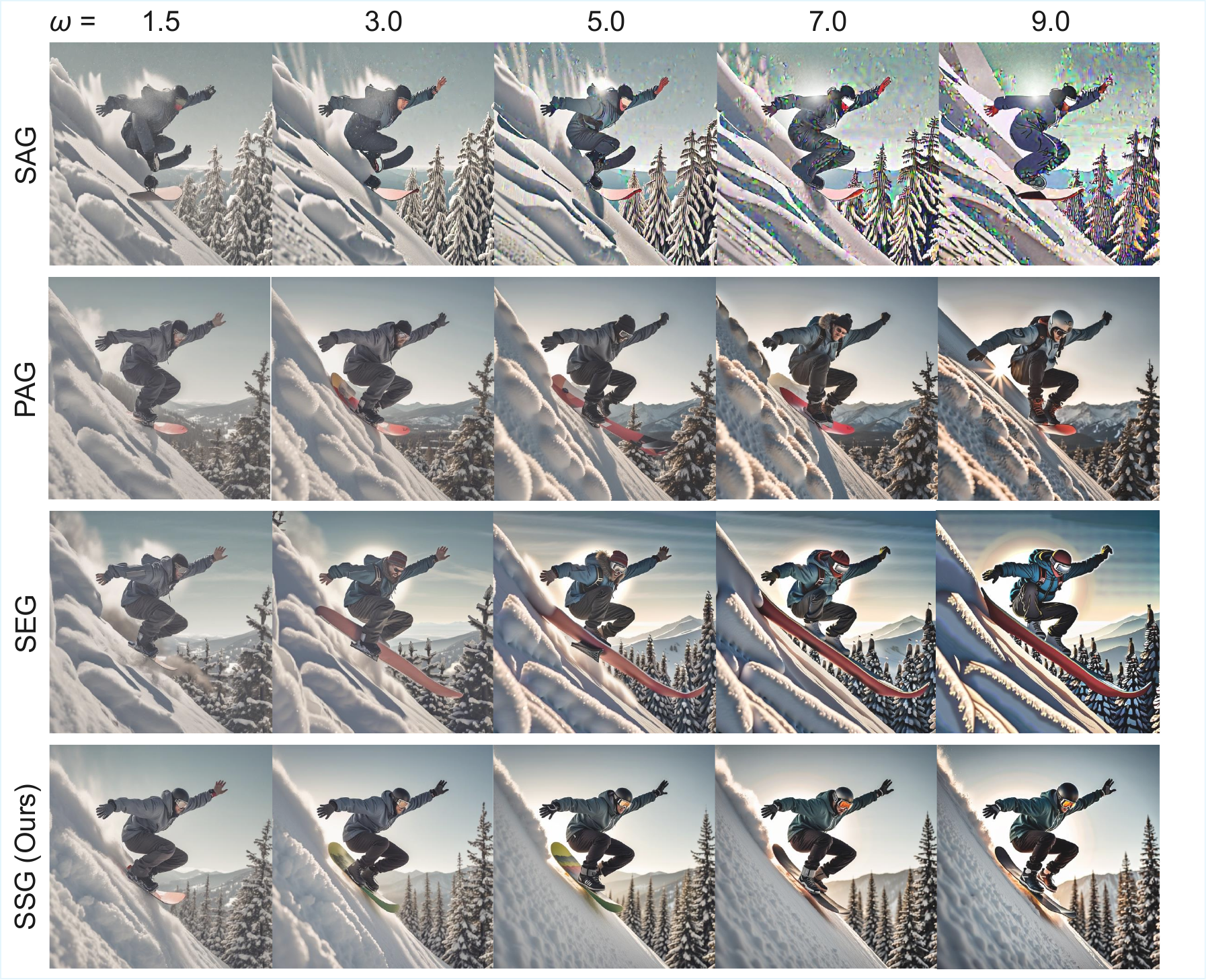} 
    \caption{\textbf{Self-Swap Guidance (SSG) generates higher-fidelity images over a wider range of guidance scale.} In contrast, existing methods~\cite{sag, pag, seg} suffer from poor details at lower guidance scale, or noise, oversaturation, and oversimplified details at higher guidance scale.}
    \label{fig:scale_sensitivity}
    \vspace{-4mm}
\end{figure} 

Diffusion models (DMs)~\cite{ddpm, ddim, sohl2015dm, song2019dm, song2020dm, sd1p5, sdxl} have emerged as a dominant paradigm in generative modelling, achieving remarkable success in image~\cite{sd1p5, sdxl, saharia2022photorealistic, epstein2023diffusion, dit, chen2024pixart, esser2024scaling}, video~\cite{vdm, svd, cogvidx, gupta2024photorealistic}, and 3D/4D content~\cite{dreamfusion, prolificdreamer,  magic3d, singer2023text, 4real} synthesis problems. A crucial factor behind their success is the use of sampling guidance~\cite{cg, cfg, cfg++}—an inference-time mechanism that steers the iterative denoising process towards higher-quality and more semantically meaningful outputs. Sampling guidance operates by providing a negative reference signal that the sampling process should avoid, thereby nudging the model towards more desirable regions of the data distribution. Amongst these methods, Classifier-Free Guidance (CFG)~\cite{cfg} is perhaps the most well-known, and is widely adopted in state-of-the-art diffusion models~\cite{sd1p5, sdxl}. By providing an empty text prompt to the model, CFG obtains an unconditional generation and uses it as a negative reference, guiding the model towards outputs that are both visually appealing and more aligned with textual instructions. 

\begin{figure*}[th]
    \centering
\includegraphics[width=0.80\textwidth]{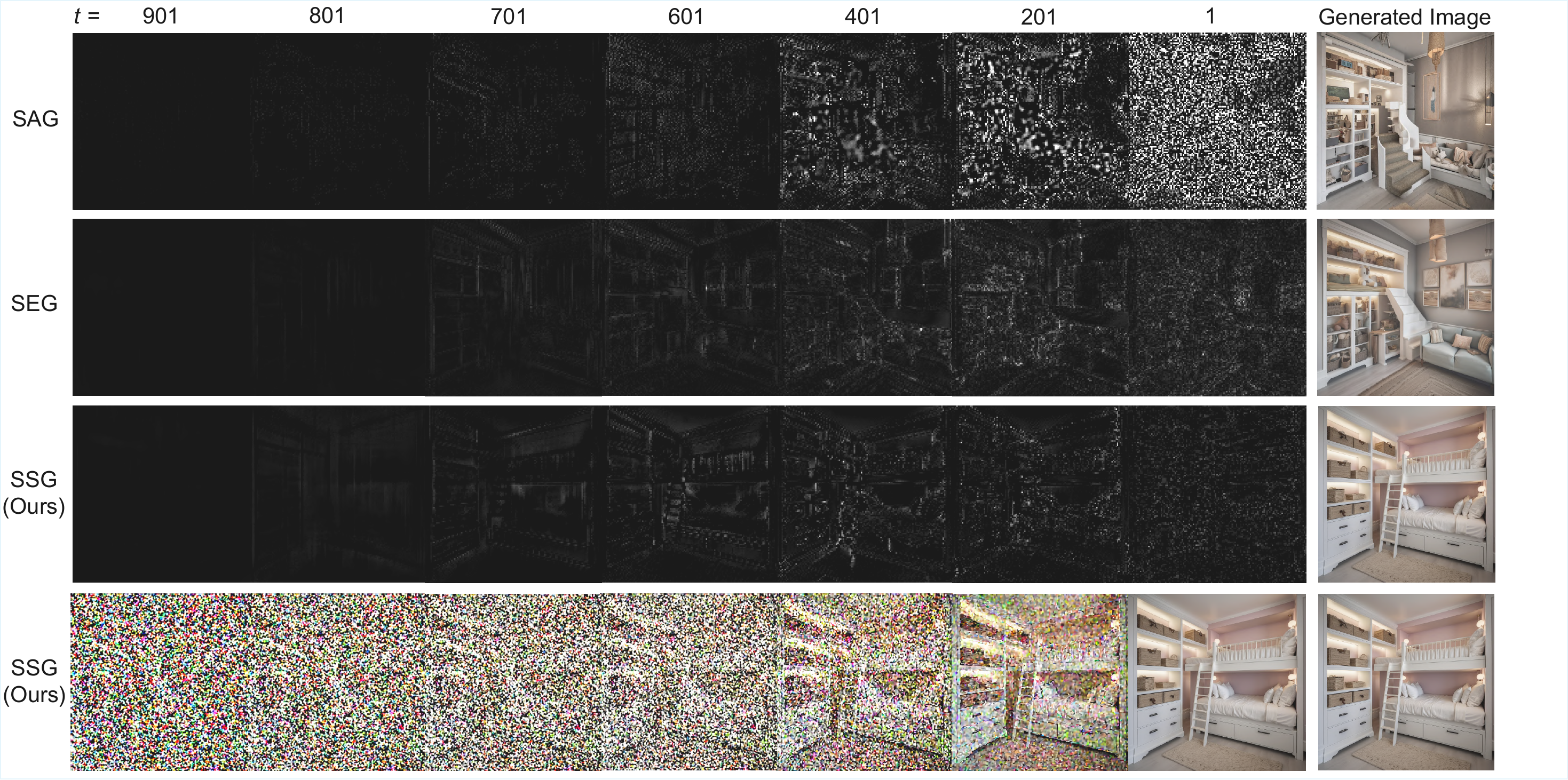} 
    \caption{\textbf{Visualisations of guidance patterns and the iteratively denoised images across different timesteps.} The text prompt used is ``\textit{A loft bed with a dresser underneath it}''.}
    \label{fig:vis_delta} \vspace{-3mm}
\end{figure*}

Despite their effectiveness, CFG and other early guidance approaches~\cite{cg, cfg, glide} require external conditioning information such as text~\cite{glide, cfg} or class labels~\cite{cg}. This dependency prevents their use in unconditional generation settings, such as solving inverse problems~\cite{dps, stsl, daps}. Additionally, CFG’s reliance on specialised training procedures (\eg, random text dropout) and tendency to produce oversaturated or less diverse samples at high guidance scales further limit its general applicability~\cite{autoguidance, apg, saharia2022photorealistic}.

To address these limitations of CFG, recent works~\cite{autoguidance, sag, pag, seg, tsg} have explored \emph{condition-free} guidance strategies to unlock the benefits of inference-time guidance for universal unconditional and conditional generation.
Without relying on external conditions like text, these methods manually perturb the model’s internal forward process during inference, effectively simulating a weaker model branch that serves as a self-contained negative reference. 
In these recent methods, perturbations are introduced in the form of noise added to the inputs~\cite{sag, tsg} or attention maps~\cite{pag, seg}. For example, SAG~\cite{sag} and TSG~\cite{tsg} corrupt pixels or timestep embeddings with  Gaussian noise; SEG~\cite{seg} and PAG~\cite{pag} introduce manual noise to perturb the attention maps.

Existing condition-free guidance methods~\cite{sag, pag, seg, tsg} introduce perturbations in a global and often indiscriminate manner. By applying the same perturbation patterns across the entire input or feature space, all layers, or all iterations, they treat the network’s diverse internal representations uniformly, disregarding the distinct semantic and structural roles played by different regions, channels, layers, and timesteps~\cite{gupta2024pre, meng2024not, kim2025revelio}. Consequently, these perturbations can either be too weak—failing to sufficiently disrupt key features—or overly strong, causing irrecoverable distortions and loss of details~\cite{apg, seg, tag}, illustrated in Figure~\ref{fig:vis_delta}. As a result, existing methods work in a relative narrow range of guidance scale values which requires careful parameter tuning. At lower guidance scale, they suffer from poor details and local structures, whereas at higher scale, they tend to generate noisy, oversaturated or oversimplified images~\cite{autoguidance, sag, pag, seg}, as demonstrated in Figure~\ref{fig:scale_sensitivity}.

In this paper, we propose a simple yet effective alternative: Self-Swap Guidance (SSG). Instead of injecting external noise, SSG introduces perturbation by swapping a portion of token latents within the model’s intermediate representation space. At token and channel granularity, this modification encourages the model to focus on fine-grained details such as textures and edges as well as global coherence. 
It yields a rich and localised degradation that disrupts both semantic and structural consistency without introducing destructive noise.
By selectively and dynamically swapping the most semantically dissimilar spatial-wise or channel-wise token latents across layers and timesteps, SSG generates a weakened model branch that is more informative and controllable, which contributes to a more effective sampling guidance trajectory.

\begin{figure*}[t]
    \centering
    \includegraphics[width=0.72\textwidth]{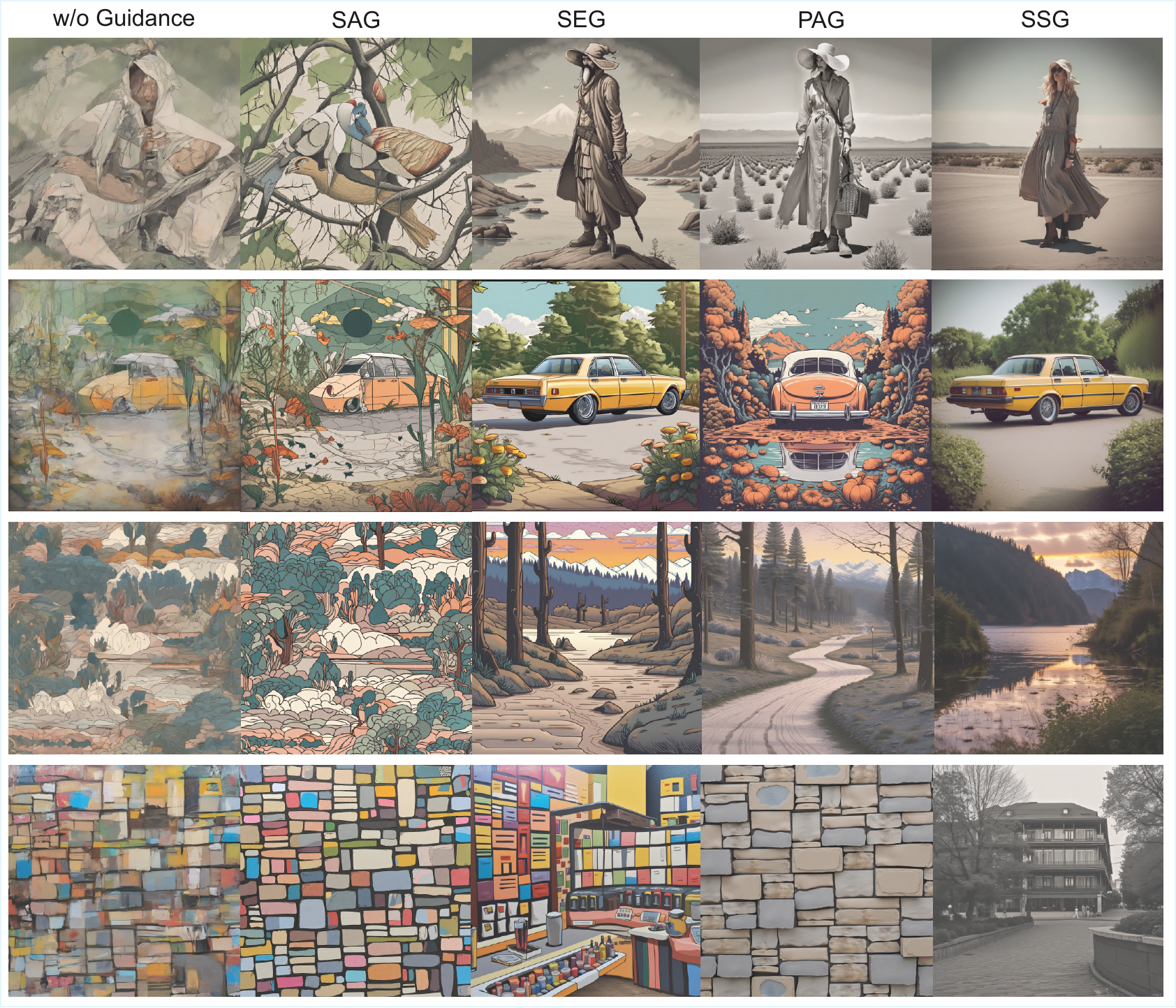} 
    \caption{\textbf{Qualitative comparison of unconditional image generation by SDXL.}}
    \label{fig:vis_uncond} \vspace{-4mm}
\end{figure*}

We evaluate SSG on both conditional and unconditional image generation tasks using SDXL~\cite{sdxl} and SD1.5~\cite{sd1p5}. In both tasks, SSG consistently produces more realistic and more faithful images compared to previous condition-free guidance methods such as SAG~\cite{sag}, PAG~\cite{pag}, and SEG~\cite{seg}, when evaluated against MS-COCO 2014/2017~\cite{coco2014} and ImageNet~\cite{imagenet} data (comparisons with the concurrent work of TPG~\cite{tpg} are presented in the supplementary material).
It exhibits more stable behaviour and maintains better visual realism across a wider range of guidance scales, as shown in Figure~\ref{fig:scale_sensitivity}.
SSG naturally integrates into standard diffusion pipelines as a lightweight plug-in, without the need for any retraining or architectural modification. Under conditional generation set-ups, it is also  compatible with CFG~\cite{cfg}, allowing joint or stand-alone usage depending on the desired trade-off amongst fidelity, diversity, and alignment with prompts. 
\section{Related Work}
\paragraph{Diffusion models.}
Diffusion models (DMs)~\cite{sohl2015dm, song2019dm, song2020dm} have emerged as a powerful class of generative models that iteratively refine noise into coherent data samples through a denoising process. Amongst the earliest variants, Denoising Diffusion Probabilistic Models (DDPM)~\cite{ddpm} established the foundation by formulating generation as a Markov chain of denoising steps. Subsequently, Denoising Diffusion Implicit Models (DDIM)~\cite{ddim} introduced a deterministic formulation that allows for faster sampling without sacrificing sample quality. Building upon these ideas, Latent Diffusion Models (LDMs)~\cite{sd1p5} perform diffusion within a learned latent space rather than directly in the raw pixel space, greatly reducing computational requirements while maintaining high perceptual fidelity. In this work, we adopt SD1.5~\cite{sdxl} and SDXL~\cite{sdxl} as our backbone models for text-conditioned and unconditional image generation. They are instances of LDMs and allow efficient high-resolution and photorealistic image synthesis.

\paragraph{Sampling guidance for diffusion models.}
Sampling guidance methods~\cite{glide, cg, cfg, cfg++} are essential to further enhancing the generation quality of diffusion models during inference. They steer the denoising trajectory of diffusion models away from manually degraded samples, obtained via input or model perturbations, and towards the high-quality data manifold.
Behind the success of state-of-the-art diffusion models~\cite{sd1p5, sdxl}, Classifier-Free Guidance (CFG) uses unconditional generation as the degraded branch to obtain higher quality and better prompt-aligned images. However, CFG requires specialised training procedures, and tends to generate oversaturated or oversimplified samples in reduced diversity~\cite{cads, interval, autoguidance}.
Worse, CFG and other early guidance methods~\cite{cg, glide, apg} rely on external conditions such as text~\cite{glide, cfg, apg} or class~\cite{cg, apg}, making them unusable in unconditional generation settings~\cite{autoguidance, dps, stsl, daps}.

\begin{table*}[t]
\centering
\begin{minipage}{0.48\textwidth}
\centering
\setlength{\tabcolsep}{4pt}
\resizebox{\textwidth}{!}{
\begin{tabular}{ccccccc}
\toprule
\textbf{Method} & \textbf{\#Steps} & \textbf{FID}$\downarrow$ & \textbf{IS}$\uparrow$ & \textbf{Precision}$\uparrow$ & \textbf{Recall}$\uparrow$ & \textbf{AES}$\uparrow$  \\
\midrule
w/o guidance & 50 & 119.04 & 9.082 & 0.277 & 0.085 &  5.646  \\
SAG~\cite{sag} & 50 & 113.33 & 8.765 & \underline{0.377} & 0.184 & 5.851   \\
SEG~\cite{seg} & 50 & \underline{89.29} & 12.53 & 0.276 & \textbf{0.257} & \underline{5.939} \\
PAG~\cite{pag} & 50 & 103.72 & \underline{13.59} & 0.265 & 0.218 & 5.734   \\
\cellcolor{paleblue} \textbf{SSG (ours)} & \cellcolor{paleblue} 50 & \cellcolor{paleblue} \textbf{70.91} & \cellcolor{paleblue} \textbf{16.44} & \cellcolor{paleblue} \textbf{0.380} & \cellcolor{paleblue} \underline{0.227} &  \cellcolor{paleblue} \textbf{6.034}  \\
\bottomrule
\end{tabular}} \vspace{-0mm}
\caption{\textbf{Quantitative comparison of unconditional image generation by SDXL on MS COCO-2014.}} \label{tab:uncond_coco14}
\vspace{-3mm}
\end{minipage}
\hfill
\begin{minipage}{0.48\textwidth}
\centering
\setlength{\tabcolsep}{4pt}
\resizebox{\textwidth}{!}{
\begin{tabular}{ccccccc}
\toprule
 \textbf{Method} & \textbf{\#Steps} & \textbf{FID}$\downarrow$ & \textbf{IS}$\uparrow$ & \textbf{Precision}$\uparrow$ & \textbf{Recall}$\uparrow$ & \textbf{AES}$\uparrow$   \\
\midrule
 w/o guidance & 50 & 74.11 & 14.73 & 0.707 & 0.236 & 5.005 \\
SAG~\cite{sag} & 50 & 64.29 & 17.22 & 0.701 & \textbf{0.331} & 5.123  \\
SEG~\cite{seg} & 50 & \underline{63.41} & 17.49 & 0.696 & 0.307 & 5.074 \\
PAG~\cite{pag} & 50 & 66.92 & \textbf{20.66} & \textbf{0.722} & 0.293 & \textbf{5.383} \\
\cellcolor{paleblue} \textbf{SSG (ours)} & \cellcolor{paleblue} 50 & \cellcolor{paleblue} \textbf{63.05} & \cellcolor{paleblue} \underline{18.84} & \cellcolor{paleblue} \textbf{0.722} & \cellcolor{paleblue} \underline{0.317} & \cellcolor{paleblue} \underline{5.286} \\
\bottomrule
\end{tabular}}  \vspace{-0mm}
\caption{\textbf{Quantitative comparison of unconditional image generation by SD1.5 on ImageNet.}} \label{tab:uncond_imagenet} \vspace{-3mm}
\end{minipage} 
\end{table*}

\paragraph{Condition-free sampling guidance.}
Recent attempts~\cite{sag, pag, seg, tsg} have been made to unlock the benefits of inference-time guidance for unconditional generation. Without requiring external conditions like text, these methods degrade the generation by perturbing the diffusion model itself. For instance, SAG~\cite{sag} and TSG~\cite{tsg} add noise to the input image and the timestep embedding, respectively, whereas PAG~\cite{pag} and SEG~\cite{seg} corrupt the intermediate attention maps.
Nonetheless, these methods lack granularity in their perturbations. By applying corruption universally to the input~\cite{sag, tsg} or attention~\cite{pag, seg} space, they either fail to sufficiently disrupt key structures and features, yielding poor fine details, or excessively distort the image, causing undesirable effects such as oversaturation or oversimplification~\cite{autoguidance, seg, tag}.
In this paper, we introduce a simple condition-free guidance method that selectively perturbs structures and semantics without broadly or excessively disturbing the forward pass. It yields high-quality images across a wider range of perturbation strengths and offers finer control over the quality–diversity trade-off.
\section{Preliminaries}
\paragraph{Diffusion models.}
Diffusion models~\cite{ddpm, ddim, sohl2015dm, song2019dm, song2020dm, sd1p5, sdxl} are a class of generative models that learn to synthesise data by reversing a gradual noising process. Starting from clean samples, noise is incrementally added following a stochastic differential equation (SDE) that describes the forward diffusion process~\cite{song2020dm}:
\begin{equation}
    dx = -\frac{\beta(t)}{2}x\,dt + \sqrt{\beta(t)}\,dw,
\end{equation}
where $\beta(t)$ represents a time-dependent noise schedule and $w$ denotes the standard Wiener process. The corresponding reverse-time dynamics, which transform noise back into structured data, can be written as:
\begin{equation}
    dx = \Big[-\frac{\beta(t)}{2}x - \beta(t)\nabla_x \log p_t(x)\Big]dt + \sqrt{\beta(t)}\,d\bar{w},
\end{equation}
where $d\bar{w}$ denotes the Wiener process evolving backward in time. $\nabla_x \log p_t(x)$ is the score function, representing the gradient of the log-density at time $t$. In practice, this score function is approximated by a neural network $s_\theta(x_t)$, which is trained with the following denoising score matching objective~\cite{dsm, song2019dm}:
\begin{equation}
\begin{aligned}
    \theta^* = \arg\min_\theta \mathbb{E}_t \Big\{ \lambda(t)\,
    \mathbb{E}_{x_0} \mathbb{E}_{x_t|x_0} \big[ 
    \| s_\theta(x_t) \\ - \nabla_x \log p_t(x_t|x_0) \|_2^2 
    \big] 
    \Big\}.
\end{aligned}
\end{equation}

During inference, the learned score network iteratively refines a random noise vector into a coherent image by numerically solving the reverse SDE.

\paragraph{Inference-time sampling guidance.}
Classifier-Free Guidance (CFG)~\cite{cfg} steers the generated sample towards higher-quality, more text-aligned outputs with a linear extrapolation of conditional and unconditional model predictions:
\begin{equation}
    \tilde{\epsilon}_{\text{CFG}}(x_t, y) = \epsilon_{\text{cond}}(x_t, y) 
    + \omega \big( \epsilon_{\text{cond}}(x_t, y) - \epsilon_{\text{uncond}}(x_t, \varnothing) \big),
\end{equation}
\noindent where $\epsilon_{\text{uncond}}(x_t, \varnothing))$ is the unconditional noise prediction at timestep $t$, obtained by supplying the model with an empty text (\ie, $\varnothing$), and serve a `negative reference to steer away from. $\omega$ is the guidance scale that controls the trade-off of sample quality versus text alignment and image quality.
In condition-free guidance methods, the negative example is constructed by perturbing the model's forward mechanism, such as corrupting the input~\cite{sag, tsg} or attention maps~\cite{pag, seg}. Denoting this degraded prediction as $\epsilon_{\text{pert}}(x_t)$, the guidance is defined by analogy:
\begin{equation} \label{eq:guidance}
    \tilde{\epsilon}(x_t) = \epsilon_{\text{ori}}(x_t) 
    + \omega \big( \epsilon_{\text{ori}}(x_t) - \epsilon_{\text{pert}}(x_t) \big),
\end{equation} 
\noindent where $\epsilon_{\text{ori}}(x_t)$ is the clean prediction left unperturbed. Notice that external condition $y$ (\eg, text) is optional in this formulation, which means condition-free guidance may support both unconditional and conditional generation.
\begin{figure*}[t]
    \centering
    \includegraphics[width=0.90\textwidth]{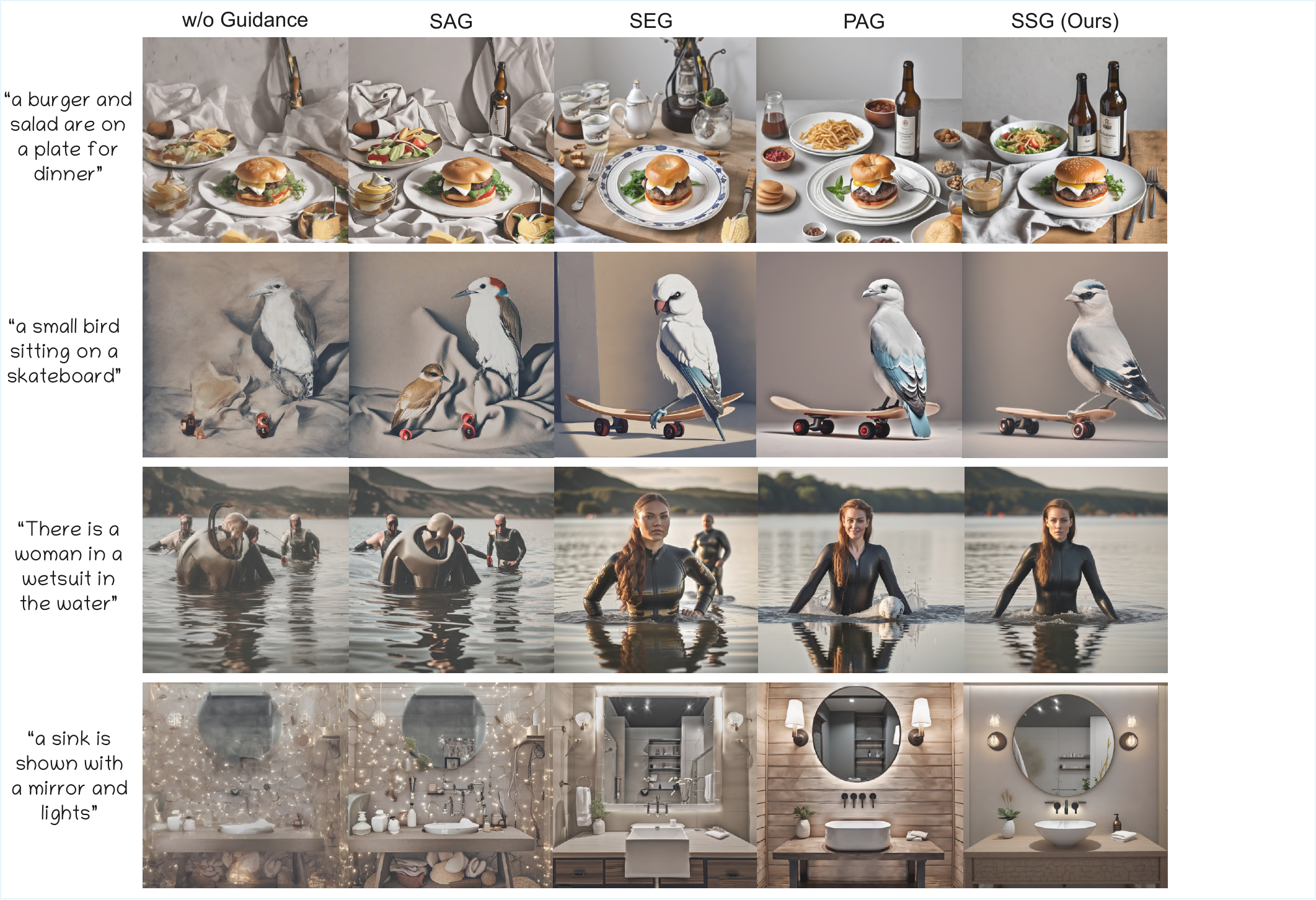} 
    \caption{\textbf{Qualitative comparison of conditional image generation by SDXL.}}
    \label{fig:vis_cond} \vspace{-3mm}
\end{figure*}

\section{SSG: Self-Swap Guidance}
\paragraph{Degrading generation with token swap.} 
The central idea of condition-free sampling guidance is to manually perturb the forward propagation process of the diffusion model to produce a \emph{weakened} version of the model's generation~\cite{autoguidance, sag, pag, seg}. Existing approaches typically introduce such perturbations in a global and less-constrained manner---for example by adding noise uniformly to inputs or attention maps across layers and  timesteps~\cite{autoguidance, sag, pag, seg}. While effective in certain contexts, these global perturbations often overlook the inherent representation diversity across network layers and the distinct temporal dynamics throughout the reverse sampling process. As a result, they often produce excessive or unrecoverable disruptions, particularly at earlier timesteps~\cite{pag, tag}. More importantly,
because these perturbations operate at coarse granularity, they provide limited control over the trade-off between perturbing and preserving quality: increasing the perturbation strength may enhance structure or texture but often introduces degradation~\cite{apg, tag}.

\paragraph{Spatial self-swap of tokens.}
To address this lack of granularity, we aim to introduce perturbations that are strong enough to effectively guide the sampling process towards better-quality outputs, yet fine-grained enough to avoid excessive or global disruption. Our key insight is that this trade-off can be better controlled at the token level. Accordingly, we propose a local and selective perturbation strategy that operates by swapping only a subset of tokens within each per-sample token feature, applied across layers and timesteps. Unlike global noise injection, this token-level swap disrupts semantic and structural relationships by modifying selected subset of semantic units, while leaving the remainder untouched. This controlled degradation produces weaker predictions that still retain essential global coherence, making the generation more resilient to oversaturation, distortion or other side-effects across a wider range of perturbation strengths.

\begin{table*}[t]
\centering
\begin{minipage}{0.48\textwidth}
\centering
\setlength{\tabcolsep}{2pt}
\resizebox{\textwidth}{!}{
\begin{tabular}{cccccccc}
\toprule
\textbf{Method} & \textbf{\#Steps} & \textbf{FID}$\downarrow$ & \textbf{CLIP}$\uparrow$ & \textbf{IS}$\uparrow$ & \textbf{AES}$\uparrow$ & \textbf{PickScore}$\uparrow$ & \textbf{IR}$\uparrow$  \\
\midrule
w/o guidance & 50 & 45.09 & 0.281 & 21.31 & 5.671 & 20.20 & -0.847 \\
SAG~\cite{sag} & 50 & 34.14 & 0.295 & 22.95 & 5.745 & 20.64 & -0.487 \\
SEG~\cite{seg} & 50 & 28.55 & 0.302 & 27.16 & \underline{5.894} & 21.38 & -0.0160 \\
PAG~\cite{pag} & 50 & \underline{26.55} & \underline{0.306} & \underline{29.70} & 5.820 & \underline{21.56} & \underline{-0.00318} \\
\cellcolor{paleblue} \textbf{SSG (ours)} & \cellcolor{paleblue} 50 & \cellcolor{paleblue} \textbf{21.73} & \cellcolor{paleblue} \textbf{0.313}  & \cellcolor{paleblue} \textbf{34.63} & \cellcolor{paleblue} \textbf{5.902} &  \cellcolor{paleblue} \textbf{22.17} & \cellcolor{paleblue}  \textbf{0.276} \\ 
\bottomrule
\end{tabular}}
\caption{\textbf{Quantitative comparison of conditional image generation by SDXL on MS-COCO 2014.}} \label{tab:cond_coco14}  \vspace{-0mm}
\end{minipage} \vspace{-0mm}
\hfill
\begin{minipage}{0.48\textwidth}
\centering
\setlength{\tabcolsep}{2pt}
\resizebox{\textwidth}{!}{
\begin{tabular}{cccccccc}
\toprule
\textbf{Method} & \textbf{\#Steps} & \textbf{FID}$\downarrow$ & \textbf{CLIP}$\uparrow$ & \textbf{IS}$\uparrow$ & \textbf{AES}$\uparrow$ & \textbf{PickScore}$\uparrow$ & \textbf{IR}$\uparrow$ \\
\midrule
w/o guidance & 50 & 54.93 & 0.280 & 21.00 & 5.660 & 20.17 & -0.870 \\
SAG~\cite{sag} & 50 & 43.76 & 0.293 & 23.08 & 5.745 & 20.60 & -0.515 \\
SEG~\cite{seg} & 50 & 38.66 & 0.301 & 27.70 & \underline{5.881} & 21.35 & -0.0575 \\
PAG~\cite{pag} & 50 & \underline{36.94} & \underline{0.306} & \underline{29.03} & 5.817 & \underline{21.54} & \underline{-0.0145} \\
\cellcolor{paleblue} \textbf{SSG (ours)}& \cellcolor{paleblue} 50 & \cellcolor{paleblue} \textbf{31.92} & \cellcolor{paleblue} \textbf{0.312} & \cellcolor{paleblue} \textbf{34.15} & \cellcolor{paleblue} \textbf{5.890} & \cellcolor{paleblue} \textbf{22.14} & \cellcolor{paleblue} \textbf{0.253} \\
\bottomrule
\end{tabular}}
\caption{\textbf{Quantitative comparison of conditional image generation by SDXL on MS-COCO 2017.}}
\label{tab:cond_coco17}
\end{minipage} \vspace{-3mm}
\end{table*}

\paragraph{Channel self-swap of tokens.}
Building on spatial self-swap, we further extend our perturbation strategy to the channel dimension through channel swaps of token features. Whereas spatial swaps primarily influence structural and geometric coherence, channel swaps perturb tokens along their channel embeddings, encouraging the model to refine subtle feature correlations such as textures, materials, and global appearance attributes. By jointly leveraging spatial and channel swaps, the model receives a balanced, fine-grained perturbation that strengthens both local detail and overall photorealism.

\paragraph{Adversarial token swap.} 
Another important insight of our approach is that swapping semantically dissimilar tokens—--rather than randomly selected ones--—leads to better generation. This strategy, inspired by adversarial analysis of vision transformers and generative models~\cite{zhao2023generative, ming2024boosting}, produces a more effective weakening of the model without requiring widespread perturbation. 
The implementation of the proposed token self-swap operations is simple and straightforward.
Specifically, for spatial self-swap, given a batch of token embeddings $\mathbf{X} \in \mathbb{R}^{B \times T \times D}$, we first normalise all token vectors along the feature dimension and compute the cosine similarity between pairs of tokens across spatial positions. 
The $N$ token pairs with the lowest similarity scores are selected, where $N$ is determined by a predefined swap ratio $r$. The indices of these pairs are used to construct a permutation mapping that exchanges the corresponding tokens in a  parallel manner. This design spatially swaps the most semantically dissimilar token pairs for each instance. Channel self-swap is implemented likewise, in a reciprocal manner.

\paragraph{Integration into diffusion models.} 
To integrate SSG into existing diffusion models, we maintain two parallel branches during forward propagation. The original branch is left unmodified, producing $ \epsilon_{\text{ori}}$ in Equation~\ref{eq:guidance}, while the degradation branch applies the proposed token swaps and produces $ \epsilon_{\text{pert}}$. Intermediate predictions from both branches are concatenated at each stage and each timestep, allowing efficient parallel processing with minimal computational overhead. We apply token self-swap operations at the beginning of each transformer blocks~\cite{transformer} and before residual shortcuts~\cite{resnet} for maximised disruptive effect.
\section{Experiments}
\subsection{Experimental settings}

\paragraph{Models.} 
We evaluate the proposed method on two popular open-sourced diffusion models for image synthesis, namely Stable Diffusion v1.5 (SD1.5)~\cite{sd1p5} and Stable Diffusion XL (SDXL)~\cite{sdxl}. We use the publicly available pretrained weights from their official repository, and adopt default inference configurations unless otherwise specified.

\paragraph{Dataset and metrics.} 
We use the MS-COCO 2014, MS-COCO 2017~\cite{coco2014}, and ImageNet~\cite{imagenet} validation sets as the reference data to evaluate the quality of generated images. For conditional generation, we sample 30k captions of MS-COCO 2014 validation and 5k of MS-COCO 2017 validation images as the text conditions. For unconditional generation, we evaluate 30k images on MS-COCO 2014 validation and 50k images on ImageNet validation data. 
The primary evaluation metrics are Fr\'{e}chet Inception Distance (FID)~\cite{fid}, Aesthetic Score (AES)~\cite{aes}, PickScore~\cite{pickscore}, and ImageReward (IR)~\cite{imagereward} for evaluating image fidelity and aesthetics, and Inception Score (IS)~\cite{is} for diversity.
We also use CLIP Score~\cite{clipscore} to measure prompt alignment in conditional generation, and Improved Precision and Recall~\cite{pr} in unconditional generation.

\paragraph{Implementation.} Our method is implemented in PyTorch using the \texttt{diffuser}~\cite{diffuser} library.  All guidance methods (\ie, PAG~\cite{pag}, SEG~\cite{seg}, CFG~\cite{cfg}, SSG) adopt the Euler discrete scheduler~\cite{edm}, with the exception of SAG~\cite{sag}, for which only DDIM~\cite{ddim} is supported. 
Experiments are conducted on NVIDIA Tesla V100 GPUs. Further details are provided in the supplementary material.

\subsection{Main Results}
We present the main qualitative and quantitative evaluation results for SSG and other existing methods. The best quantitative result under each metric is marked in bold, and second best underlined. For conditional generation, we provide the prompt used to generate each set of images. For unconditional generation, each set uses the same random seed.

\paragraph{Unconditional generation with SSG.} Table~\ref{tab:uncond_coco14} provides a comprehensive quantitative comparison of various condition-free guidance methods on MS-COCO 2014 data. 
On SDXL, SSG consistently delivers the best performance across all metrics, leading previous methods by considerable margins. Compared to the vanilla baseline without inference-time guidance, SSG dramatically improves FID from 119.04 to 70.91 and Inception Score from 9.08 to 16.44. Compared to recent guidance methods such as PAG and SEG, it achieves the best results overall, including the highest AES score of 6.034. Table~\ref{tab:uncond_imagenet} reports results of SD1.5 on ImageNet data, where SSG achieves the best FID and overall performance. We postulate that the smaller performance gaps amongst different methods is potentially due to the lower generation quality of SD1.5, and the limited quality of the ImageNet validation set.
Qualitatively, as illustrated in Figure~\ref{fig:vis_uncond}, existing methods tend to generate non-photorealistic content and even repetitive patterns or textures under unconditional generation. In contrast, SSG is less prone to generating low-quality images and more likely to render realistic textures and coherent layouts. 

\paragraph{Conditional generation with SSG.} We present quantitative results on the SDXL model on MS-COCO 2014 evaluation in Table~\ref{tab:cond_coco14}. SSG leads to drastic improvements over the vanilla baseline across all metrics.
For example, it achieves more than $2\times$ reduction in FID and $1.5\times$ improvement in Inception Score. SSG consistently leads all previous guidance methods, with particularly large improvements on ImageReward.
We also present quantitative results using MS-COCO 2017 validation samples in Table~\ref{tab:cond_coco17}. As shown, SSG maintains its advantage across all metrics, further validating its effectiveness. Qualitative comparisons in Figure~\ref{fig:vis_cond} show that SSG is more likely to generate high-fidelity images that are both more photorealistic in terms of global coherence and local structures and textures, while also aligning more closely with the text prompts. Together, these quantitative and qualitative results corroborate the rationality and effectiveness of SSG in better guiding diffusion sampling.

\begin{figure}[t]
    \vspace{-1mm}
    \centering
    \includegraphics[width=\linewidth]{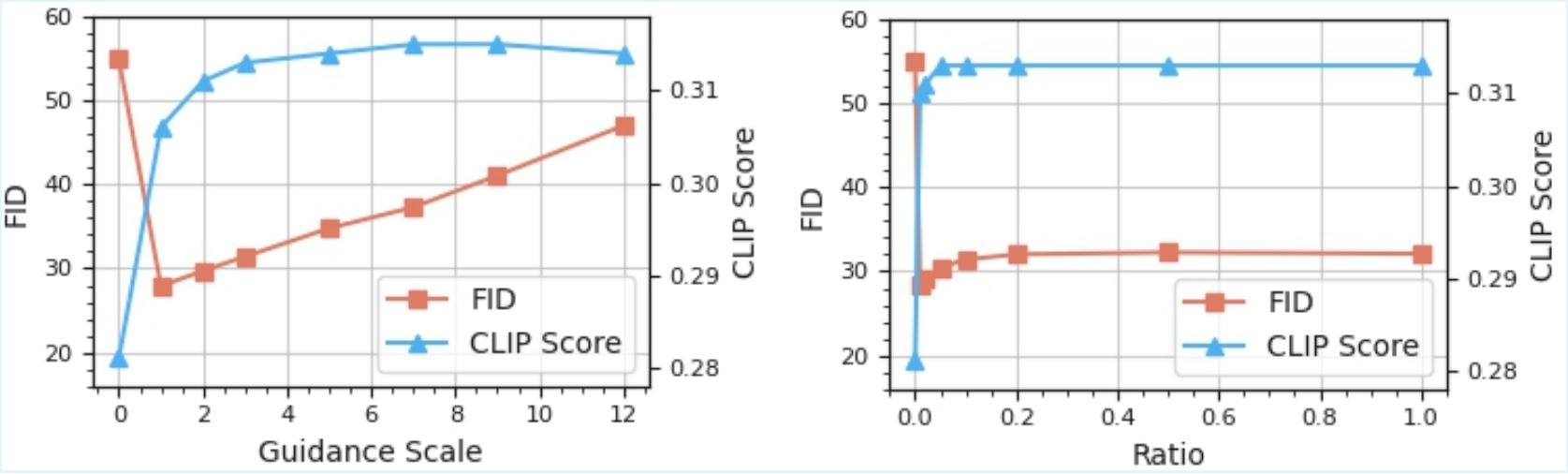}
    \vspace{-6mm}
    \caption{\textbf{Effect of varying guidance scale and swap ratio on image quality and prompt alignment.}}
    \label{fig:abl_hyperparams}
    \vspace{-2mm}
\end{figure}

\begin{figure}[t]
    \centering
    \includegraphics[width=0.47\textwidth]{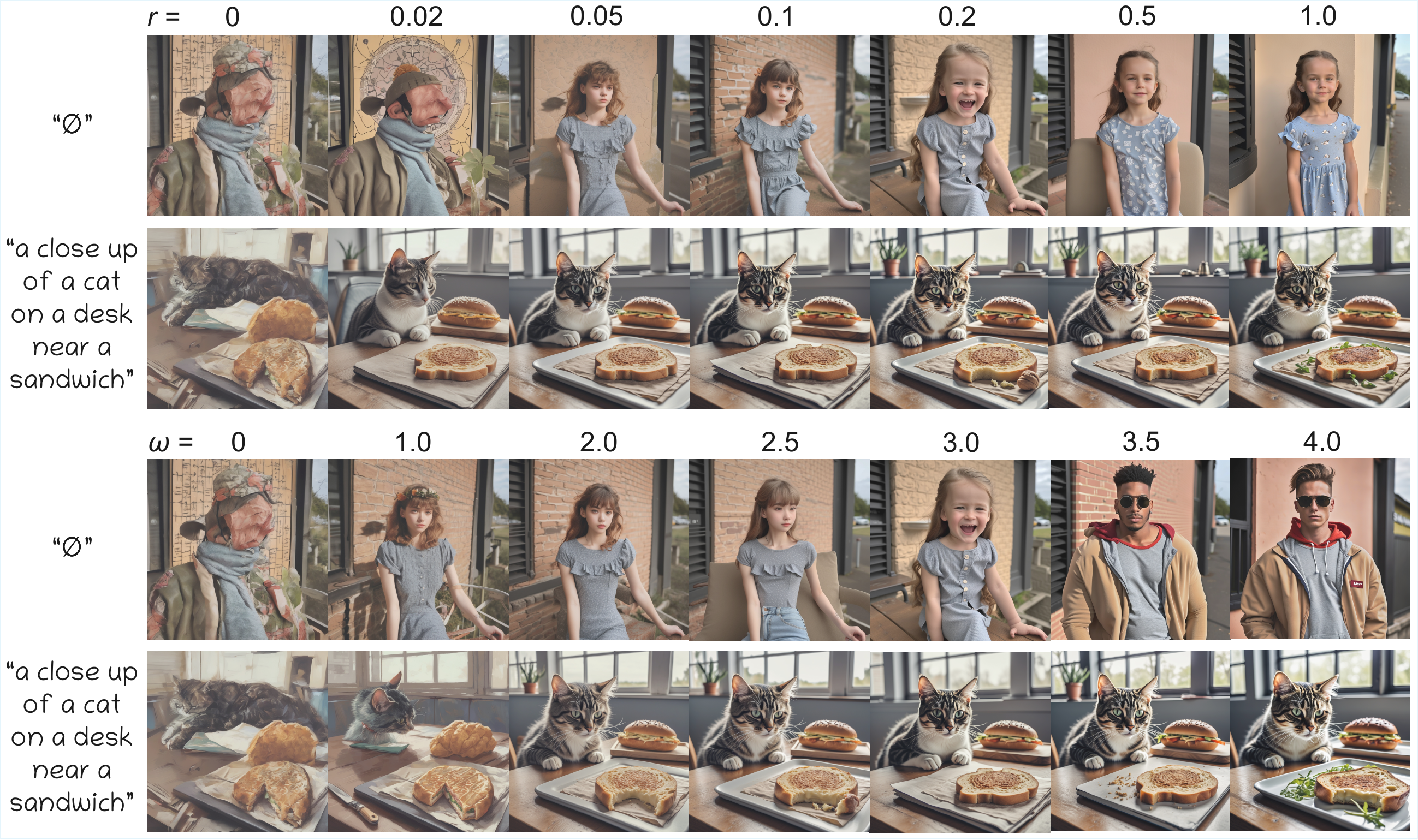} 
    \vspace{-1mm}
    \caption{\textbf{Visualising the effect of using different swap ratio and guidance scale values on generated images.}}
    \label{fig:hyperparam_sensitivity} \vspace{-3mm}
\end{figure} 

\subsection{Further Analysis} \label{sec:analysis}
We perform a set of analytical studies to gain deeper insights into the properties and rationality of our method and design.
Experiments in the section are conducted by generating 3k images using SDXL and evaluating on the MS-COCO 2014 validation data, unless otherwise specified.

\paragraph{Guidance scale.} Figure~\ref{fig:abl_hyperparams} (left) analyses the impact of larger or smaller guidance scale, $\omega$. As SSG is applied, both image quality and prompt fidelity dramatically improves, measured by FID and CLIP Score, as the guidance scale increases from zero to smaller values. Towards higher guidance values, FID gradually grows whereas CLIP Score saturates. Overall, SSG produces images of satisfactory quality over a wide range of guidance scale.
Notably, previous methods, including SAG, PAG, and SEG, are more sensitive to the guidance scale value, as illustrated in Figure~\ref{fig:scale_sensitivity}. 
At lower guidance scale, they suffer from unrealistic details (\eg, SAG). Whereas at higher guidance scale, the generated images are prone to noise (\eg, SAG), oversaturation (SAG, PAG, SEG), or oversimplified details (SAG, PAG, SEG). In contrast, SSG generates higher-quality images more consistently across different guidance scales.

\begin{table}[t]
\centering
\resizebox{0.47\textwidth}{!}{
\begin{tabular}{cccccccc}
\toprule
 \textbf{Method} &  \textbf{FID}$\downarrow$ & \textbf{CLIP}$\uparrow$ & \textbf{IS}$\uparrow$ & \textbf{AES}$\uparrow$ & \textbf{PickScore}$\uparrow$ & \textbf{IR}$\uparrow$ \\
\midrule
SAG~\cite{sag} & 43.97 & 0.295 & 22.12 & 5.756 & 20.65 & -0.483 \\ 
SEG~\cite{seg} & 38.15 & 0.303 & 26.15 & 5.888 & 21.38 & -0.0239 \\ 
PAG~\cite{pag} & 36.79 & 0.306 & 29.42 & 5.827 & 21.57 & 0.00171 \\ 
\midrule
Random & 32.28 & 0.312 & \textbf{34.67} & \textbf{5.928} & 22.14 & 0.283 \\
Similar & \textbf{28.74} & 0.309 & 32.81 & 5.826 & 21.80 & 0.110 \\
\cellcolor{paleblue} Dissimilar & \cellcolor{paleblue} 31.41 & \cellcolor{paleblue} \textbf{0.313} &  \cellcolor{paleblue} 34.44 & \cellcolor{paleblue} 5.901 & \cellcolor{paleblue} \textbf{22.18} & \cellcolor{paleblue} \textbf{0.297} \\ 
\bottomrule
\end{tabular}} \vspace{-0mm}
\caption{\textbf{Importance of adversarial token swap.} Swapping dissimilar tokens achieves the best generation quality overall. Random swap yields slightly worse results and swapping similar tokens perform worst, but they still substantially outperform.}  \label{tab:abl_ats} \vspace{-1mm}
\end{table}

\begin{figure}[t]
    \centering
    \includegraphics[width=0.48\textwidth]{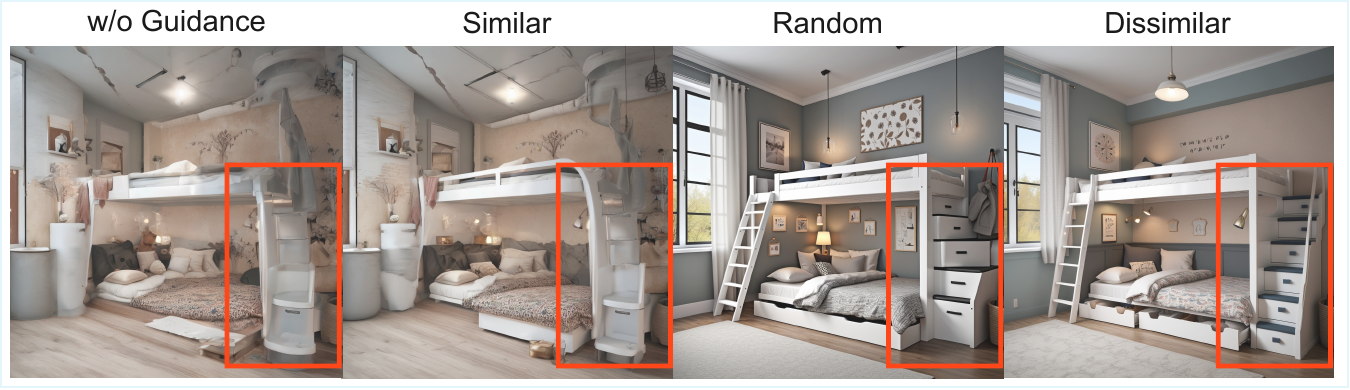} \vspace{-5mm}
    \caption{\textbf{Visualising the effect of different token swap policies.} Swapping dissimilar tokens further refines local details and global coherence compared to random swap. In contrast, swapping similar tokens leads to poor generation that resembles the vanilla diffusion model's output.}
    \label{fig:abl_ats} \vspace{-4mm}
\end{figure} 

\paragraph{Swap ratio.} The swap ratio $r$ is the ratio of swapped tokens/channels to the total number of tokens/channels. It provides a direct control over the perturbation strength in the proposed method. The right-hand plot in Figure~\ref{fig:abl_hyperparams} examines how different swap ratio impacts the generated images by SSG. Similar patterns are observed to those seen with the guidance scale. However, we notice that increasing the swap ratio from 0 to small values more drastically improve both FID and CLIP Score compared to setting guidance scale. This reveals the central role and the remarkable ability of the proposed token swap operations in guiding the samples towards higher quality. It is also noteworthy that as swap ratio further increases, the FID degradation saturates, different from the more severe degradation observed at high guidance scale, which corroborates the advantage of the proposed swap-based operation as a more fine-grained and constrained type of perturbation.

\paragraph{Adversarial token swap.}
We demonstrate the importance of swapping dissimilar token pairs to sufficiently disrupt local structures. In Table \ref{tab:abl_ats}, we consider three different policies for selecting which pairs of tokens to swap: randomly selecting $N$ pairs, selecting the $N$ most similar pairs, and our method of selecting the $N$ most dissimilar pairs. It turns out that swapping the most dissimilar pairs leads to the best overall performance the highest scores in terms of CLIP Score, PickScore, and ImageReward, whereas swapping the most similar ones performs the worst. We also provide visual examples of images generated using different swap strategies in Figure~\ref{fig:abl_ats}.
These ablation results highlight the significance of the proposed adversarial token swap. Additionally, it is surprising that even random token shuffle substantially outperforms the state-of-the-art methods of SAG and SEG -- compelling evidence to the fact that that token swap itself is a very effective form of perturbation in guiding diffusion model sampling.

\begin{table}[t]
\centering
\setlength{\tabcolsep}{4pt}
\resizebox{0.48\textwidth}{!}{
\begin{tabular}{cccccccc}
\toprule
\textbf{Spatial} & \textbf{Channel} & \textbf{FID}$\downarrow$ & \textbf{CLIP}$\uparrow$ & \textbf{IS}$\uparrow$ & \textbf{AES}$\uparrow$ & \textbf{PickScore}$\uparrow$ & \textbf{IR}$\uparrow$ \\
\midrule
 \ding{51} &  \ding{55} & 31.96 & 0.313 & 33.33 & \textbf{5.931} & 22.11 & 0.272 \\
 \ding{55} &  \ding{51}  & \textbf{31.30} & 0.313  & \textbf{34.95}  & 5.892 & 22.17 & 0.286 \\
\cellcolor{paleblue}  \ding{51} &  \cellcolor{paleblue}  \ding{51}  & \cellcolor{paleblue}  31.41 & \cellcolor{paleblue}  0.313 & \cellcolor{paleblue} 34.44 & \cellcolor{paleblue}  5.901 & \cellcolor{paleblue} \textbf{22.18} & \cellcolor{paleblue} \textbf{0.297} \\ 
\bottomrule
\end{tabular}} \vspace{-1mm}
\caption{\textbf{Ablation on two types of token swap.}}  \vspace{-0mm}
\label{tab:abl_swap}
\end{table}

\begin{table}[t]
\centering
\resizebox{0.48\textwidth}{!}{
\begin{tabular}{cccccccc}
\toprule
\textbf{SSG} & \textbf{CFG} & \textbf{FID}$\downarrow$ & \textbf{CLIP}$\uparrow$ & \textbf{IS}$\uparrow$ & \textbf{AES}$\uparrow$ & \textbf{PickScore}$\uparrow$ & \textbf{IR}$\uparrow$ \\
\midrule
\ding{55} & \ding{55} & 54.96 & 0.281 & 21.20 & 5.676 & 20.22 & -0.834 \\
\ding{51} & \ding{55} & 31.41 & 0.313 & 34.44 & \textbf{5.901} & 22.18 & 0.297\\
\cellcolor{paleblue} \ding{51} & \cellcolor{paleblue} \ding{51} & \cellcolor{paleblue} \textbf{30.82} & \cellcolor{paleblue} \textbf{0.319} & \cellcolor{paleblue} \textbf{36.37} & \cellcolor{paleblue} 5.858 & \cellcolor{paleblue} \textbf{22.44}  & \cellcolor{paleblue} \textbf{0.492} \\
\bottomrule
\end{tabular}} \vspace{-1mm}
\caption{\textbf{Compatibility of SSG with CFG.}} \label{tab:abl_cfg} \vspace{-4mm}
\end{table}

\paragraph{Effect of spatial and channel swaps.}
Table~\ref{tab:abl_swap} ablates the effects of the two token swap operations. As shown by the quantitative measurements, both strategies effectively boost generation quality, while exhibiting different characteristics: spatial swap achieves a higher AES score, whereas channel swap yields better PickScore and ImageReward, Finally, using the two jointly leads to further improved empirical results on average as well as more visually appealing images, highlighting their complementary nature.  

\paragraph{Analysis of guidance patterns.}
Figure~\ref{fig:vis_delta} visualises the guidance patterns of various guidance methods across timesteps. Here, for each denoising step $t$, we plot the guidance magnitude, obtained by taking the channel-averaged absolute value of $\omega \big( \epsilon_{\text{ori}}(x_t) - \epsilon_{\text{pert}}(x_t) \big)$ in Equation~\ref{eq:guidance}. It can be noticed that SSG exhibits strong responses to prominent edges and shapes, such as the two bedposts and the ladder, at early steps, indicating that SSG already guides the model to form and refine these structures early in the iterative generation process. In contrast, SAG and SEG show only weak or negligible responses to these shapes at the same stage. Note that early formation of such these structures is critical, as it largely determine the final image layout (see Figure~\ref{fig:vis_delta} (right)). It turns out that only SSG correctly generates the content corresponding to the text. Additionally, SSG concentrates strong guidance on fewer spatial locations than SAG and SEG, evident in later denoising steps in Figure~\ref{fig:vis_delta}.

\begin{figure}[t]
    \centering
    \includegraphics[width=0.48\textwidth]{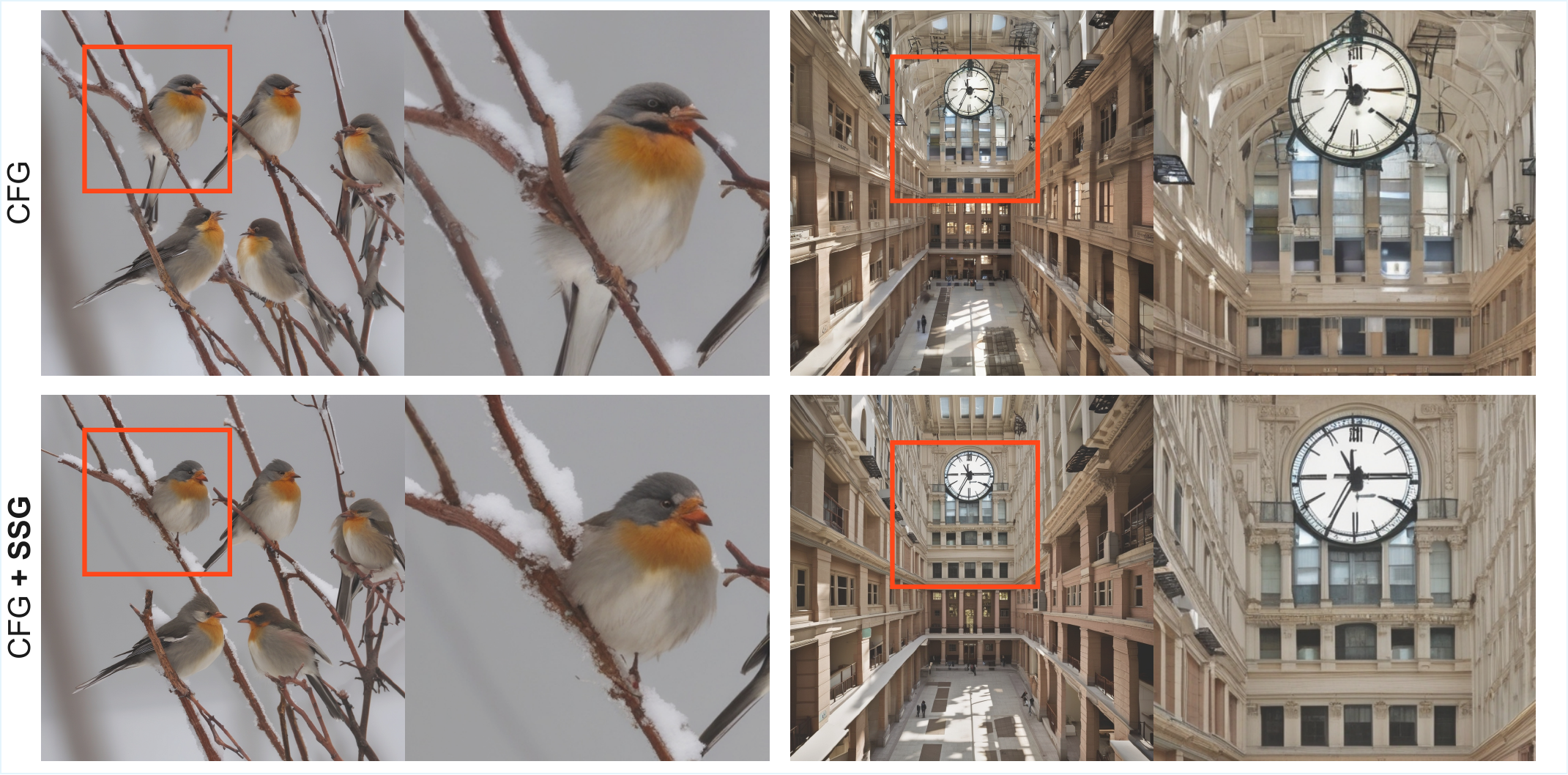} \vspace{-6mm} 
    \caption{\textbf{Compatibility between SSG and CFG.} SSG can be applied on top of CFG to further refine image quality.} \vspace{-3mm}
    \label{fig:abl_cfg}
\end{figure}

\paragraph{Compatibility with CFG.}
As SSG and CFG operate in orthogonal perturbation spaces (\ie,  token space and condition space, respectively), they can be conveniently combined to gain further improvements. To illustrate this, we apply CFG on top of SSG. Quantitative results in Table~\ref{tab:abl_cfg} show that CFG can further improve the overall image quality. 
In particular, it brings significant improvements in terms of aesthetics and prompt alignment metrics. 
Conversely, we can also expect that SSG is able to refine the intricate structural details and improve the overall fidelity of images guided by CFG. We verify this qualitatively in Figure \ref{fig:abl_cfg}, where SSG successfully enhances the fine-grained texture and shapes of substructures. These quantitative and visual evidence demonstrates that CFG and SSG complement each other and can be applied together for additional benefits.

\paragraph{Additional details, results, and analyses} including pseudo-code, further analytical studies, and more visualised examples, are presented in the supplementary material.
\section{Conclusion}
In this paper, we introduced Self-Swap Guidance (SSG), a simple, training-free, and condition-free guidance method that substantially improves the fidelity of images generated by diffusion models. 
Unlike previous approaches that perturb the entire input or attention space, SSG operates at token granularity and selectively swaps pairs of most semantically dissimilar token latents across layers and timesteps. This design provides fine-grained control over the guidance process, simultaneously disrupting local structures and global semantics without introducing destructive noise, which effectively steers the sampling process towards the generation of higher-quality images.
Extensive experiments across different diffusion models, datasets, and condition settings demonstrate that SSG consistently enhances generation quality, prompt alignment, and image diversity of diffusion models. More broadly, SSG can be readily integrated into existing diffusion pipelines as a plug-in, and is compatible with the classic CFG to allow flexible trade-offs amongst fidelity, diversity, and prompt adherence.

\paragraph{Acknowledgements.} 
This work was supported in part by NSFC (62322113, 62376156), Shanghai Municipal Science and Technology Major Project (2025SHZDZX025G15, 2021SHZDZX0102), and the Fundamental Research Funds for the Central Universities.

{
    \small
    \bibliographystyle{ieeenat_fullname}
    \bibliography{main}
}

\end{document}